\UseRawInputEncoding
\documentclass[10pt,twocolumn,letterpaper]{article}

\usepackage{cvpr}              
\usepackage{amssymb}
\usepackage{multirow}
\usepackage[dvipsnames]{xcolor}
\newcommand{\eat}[1]{}
\usepackage{wrapfig}
\usepackage{float}
\usepackage{tabularx}
\usepackage{multicol}
\usepackage{booktabs}
\usepackage{bm}
\usepackage{placeins}
\definecolor{cvprblue}{rgb}{0.21,0.49,0.74}
\usepackage[pagebackref,breaklinks,colorlinks,allcolors=cvprblue]{hyperref}

\def\paperID{13397} 
\def\confName{CVPR}
\def\confYear{2026}

\title{MiVLA: Towards Generalizable Vision-Language-Action Model with \\ Human-Robot Mutual Imitation Pre-training}
\author{%
	Zhenhan Yin$^{1}$ \quad Xuanhan Wang$^{1}$ \quad Jiahao Jiang$^{1}$ \quad Kaiyuan Deng$^2$  \quad Pengqi Chen$^2$ \quad   Shuangle Li$^2$ \\ Chong Liu$^2$ \quad Xing Xu$^1$  
	\quad Jingkuan Song$^1$ \quad Lianli Gao$^2$ \quad Heng Tao Shen$^1$ 
	\\ 
	$^1$Tongji University\\
	$^2$University of Electronic Science and Technology of China  \\
}
\eat{
\author{First Author\\
Institution1\\
Institution1 address\\
{\tt\small firstauthor@i1.org}
\and
Second Author\\
Institution2\\
First line of institution2 address\\
{\tt\small secondauthor@i2.org}
}
}
\begin{document}
\twocolumn[{
	\maketitle
	\begin{figure}[H]
		\hsize=\textwidth
		\centering
		\vspace*{-0.5in}
		\includegraphics[width=0.99\textwidth]{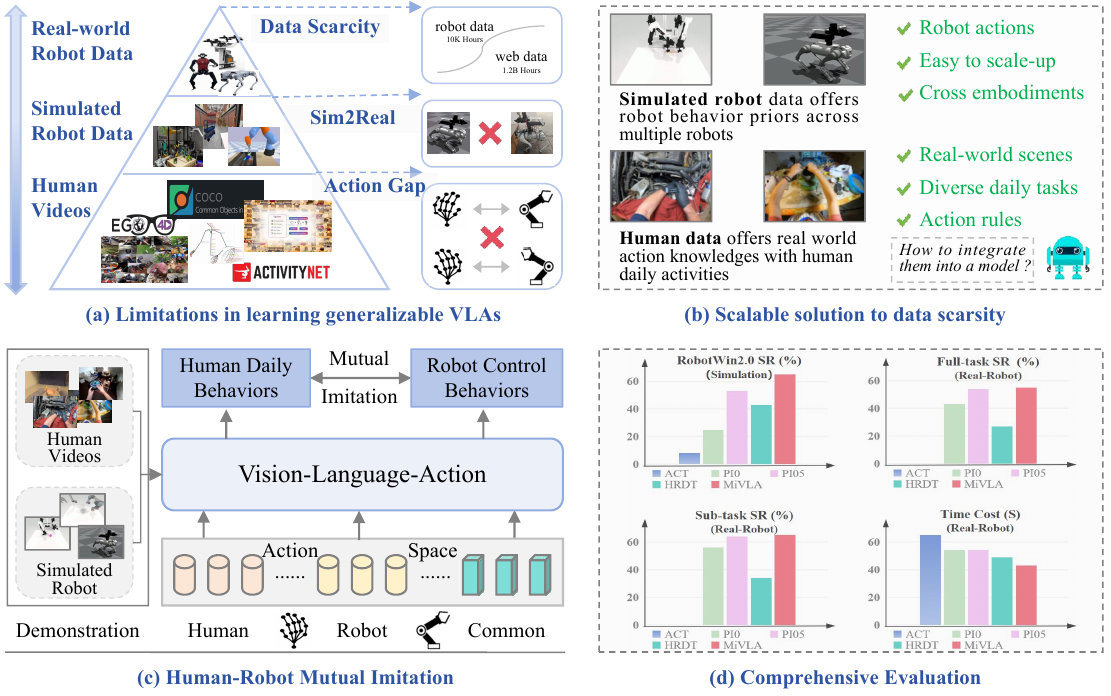}
		\caption{The motivation. (a) Vision-language-action models are generally driven by large-scale pre-training; however, their scaling is hindered by the scarcity of real-world robot demonstrations. (b) Simulated robot and human videos provide a promising alternative, as they offer not only robot action priors but also real-world behavior knowledge derived from human daily activities. (c) We propose human-robot mutual imitation to pre-train on simulation and human data, (d) achieving a generalizable model MiVLA with state-of-the-art manipulation performance on both simulation and real robot platforms.}
		\label{figure:motivation}
	\end{figure}
}]
\begin{abstract}
While leveraging abundant human videos and simulated robot data poses a scalable solution to the scarcity of real-world robot data, the generalization capability of existing vision-language-action models (VLAs) remains limited by mismatches in camera views, visual appearance, and embodiment morphologies.
To overcome this limitation, we propose MiVLA, a generalizable VLA empowered by human-robot mutual imitation pre-training, which leverages inherent behavioral similarity between human hands and robotic arms to build a foundation of strong behavioral priors for both human actions and robotic control. 
Specifically, our method utilizes kinematic rules with left/right hand coordinate systems for bidirectional alignment between human and robot action spaces. Given human or simulated robot demonstrations, MiVLA is trained to forecast behavior trajectories for one embodiment, and imitate behaviors for another one unseen in the demonstration. Based on this mutual imitation, it integrates the behavioral fidelity of real-world human data with the manipulative diversity of simulated robot data into a unified model, thereby enhancing the generalization capability for downstream tasks.
Extensive experiments conducted on both simulation and real-world platforms with three robots (ARX, PiPer and LocoMan), demonstrate that MiVLA achieves strong improved generalization capability, outperforming state-of-the-art VLAs (e.g.,  $\boldsymbol{\pi}_{0}$, $\boldsymbol{\pi}_{0.5}$ and H-RDT) by 25\% in simulation, and 14\% in real-world robot control tasks. 

\end{abstract}    
\section{Introduction}
\label{sec:intro}

\eat{
\begin{figure}[t]
	\centering
	\includegraphics[width=0.99\linewidth]{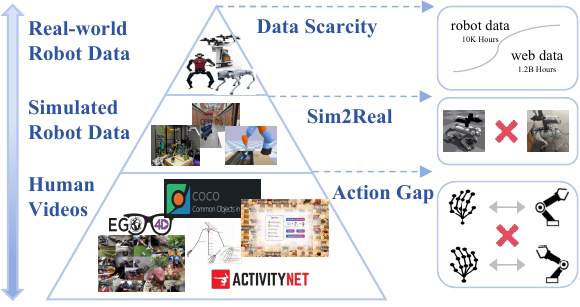}
	\caption{The challenges. }
	\vspace*{-0.1in}
	\label{fig:challenge}
\end{figure}
}
\begin{figure*}[t]
	\centering
	\includegraphics[width=0.99\linewidth]{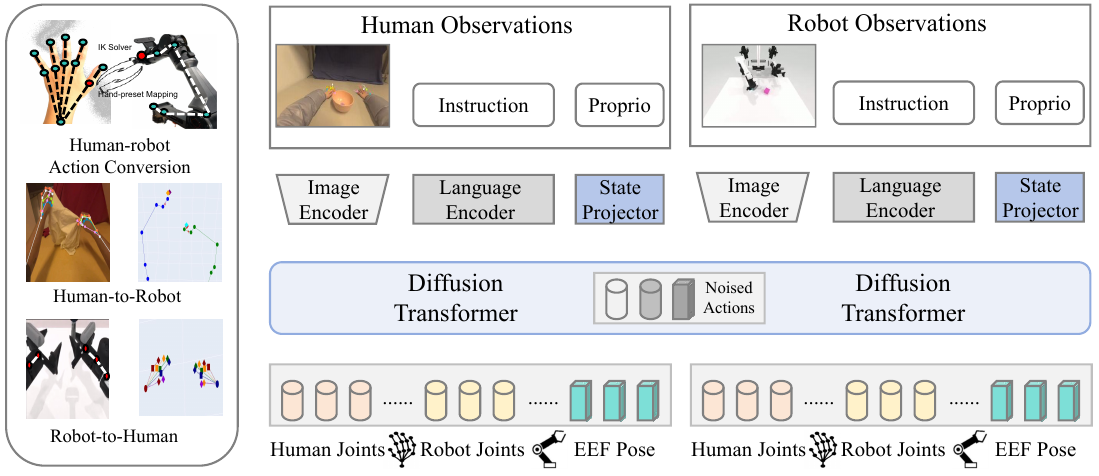}
	\caption{The overview of Proposed MiVLA. A general human-robot action mapping mechanism is introduced to bridge the gap between human-robot action space. Given a simulated robot demonstration, a VLA model is trained to predict robot action and learn to imitate robot behavior at human action space. For human demonstration, we train the same policy using human-to-robot imitation.}
	\label{figure:framework}
\end{figure*}
With the advances in general-purpose foundation models~\cite{vlm:internvl,vlm:qwen25,vlm:siglip,vfm:saipv1,vfm:dinov2}, recent years have witnessed remarkable progress in robot learning, which emerges numerous vision-language-action models (VLAs) for generalist robot policy~\cite{vla:RDT-1,vla:RT2,vla:dexvla,vla:octo,vla:pi0,vla:pi05,vla:openvla,vla:cotvla}. By training on real-world robot demonstration using imitation/reinforce learning~\cite{robot-learn:hrl,vla:pi0,robot-learn:humanplus}, VLAs successfully establish diverse multi-modal mappings from robot observations with human instructions to robot actions. 
Despite impressive success, training VLAs to master manipulative skills for flexible robot control in real-world scenarios remains challenging. This is primarily due to the scarcity of real-world robot data, which manifests in two dimensions: not only the prohibitive cost and time required to collect data at internet scale, but also the inherent difficulty in encompassing the diversity of open-world environments that we expect robots operate in. Consequently, the generalization capability and versatility of existing VLAs are significantly constrained.

To overcome the scarcity of real-world robot data, many efforts have been made to seek alternatives such as simulated robot data~\cite{simulation_data:robotwin1, simulation_data:robotwin2,simulation_data:libero,simulation_data:maniskill} and human videos~\cite{human_data:egodex,human_data:ego4d,dataset:coco,dataset:activitynet}, as it is demonstrated that the former offers behavior priors for robot control while the latter provides extensive coverage of real-world tasks and scenes. However, it poses a critical barrier to the effective transfer of those priors to real-world robot manipulation, since simulated robot data suffers from typical Sim2Real gap and human data is challenged by morphological discrepancies.
This motivates us to rethink significant questions that are less studied: \textit{\textbf{Can the complementary priors from simulated data and human data be effectively unified within a single model to create a generalizable VLA without real-world robot data? If so, what is the path to its realization?}}

In this work, we seek to answer questions by proposing MiVLA, a vision-language-action model empowered by human-robot mutual imitation pre-training, which leverages simulated robot data and human data to enable the acquisition of strong behavior priors for real-world robot control. 
Rather than directly learning multi-modal mappings from vision and language contexts to robot action space, our method first performs cross-embodiment action generation, where labeled precise actions of one embodiment are transformed into unlabeled actions for another unseen embodiment. Specifically, we utilize kinematic rules with left/right hand coordinate systems for bidirectional alignment between human and robot action spaces.  
In this way, the difference between human and robot action spaces can be approximated via bidirectional geometric transforms, and each demonstration can be equipped with complementary actions across multiple embodiments. Given human or simulated robot demonstrations with complementary actions, we pre-train a VLA to forecast behavior trajectories for one embodiment, and imitate behaviors for another one unseen in the demonstration. Based on this mutual imitation, the pre-trained VLA which is called MiVLA, integrates the behavioral fidelity of real-world human data with the manipulative diversity of simulated robot data into a unified model, thereby exhibiting strong generalization capability for downstream tasks.
In summary, the key contributions of this paper are three fold:
\begin{itemize}
	\itemsep0em 
	\item We propose a generalizable robot model termed MiVLA, which is empowered by novel mutual imitation pre-training. It incorporates not only real-world behavioral priors in human data but also manipulative diversity in simulated robot data into a unified model. 
	\item We introduce a method of bidirectional human-robot action space conversion by adopting kinematic rules with left/right hand coordinate alignment mechanism for cross-embodiment learning. 
	\item We conduct extensive experiments on both simulation environment and real robot platforms with three embodiments (PiPer, ARX, LocoMan), demonstrating mutual imitation using simulated robot and human data significantly improves generalization capability of VLA. Specifically, proposed MiVLA surpasses existing VLAs across multiple settings, achieving improved manipulation success rate in simulation tasks by 25\%, and real robot control tasks by 14\%.	
\end{itemize}

\section{Related Work}
\label{sec:related_work}

\noindent\textbf{Vision-Language-Action Models (VLAs).}
Developing a generalist robot model remains a long-standing goal in robotics. It requires not only a deep understanding of the scene but also mastering flexible robot control for interaction with the physical world~\cite{robot-perception:RoboSpatial,robot-learn:RoboPEPP}. Inspired by tremendous progress in large vision-language models (VLMs)~\cite{vlm:internvl, vlm:qwen25, vlm:paligemma2}, recent works~\cite{vla:dexvla,vla:pi0,vla:pi05,vla:egovla,vla:cotvla} have attempted to integrate them into robot control systems, deriving numerous foundational robot policies, i.e., vision-language-action models. By pre-training on robot demonstrations, VLAs learn to establish multi-modal mappings from robot observations (e.g., camera views, instructions and proprioception) to robot actions. Early works such as OpenVLA~\cite{vla:openvla} and RT-2~\cite{vla:RT2} adopt transformer-based architectures to predict discrete action tokens in an autoregressive manner. Recent efforts like $\mathbf{\mathnormal{\Pi}_0}$~\cite{vla:pi0}, DexVLA~\cite{vla:dexvla} and H-RDT~\cite{vla:hrdt} turn to diffusion-based action prediction, since discrete action tokenization brings a limitation in continuous robot control. However, achieving high-performance VLAs relies on large-scale demonstrations with right pre-training recipes~\cite{vla:pi05}. Although impressive, existing VLAs remain less generalizable and lack robustness, as they suffer from scarcity of real-world robot demonstration. In a different line, this work explores an alternative way for generalizable VLAs by leveraging abundant simulated robot data and real-world human videos. 

\noindent\textbf{Human-Robot Knowledge Transfer.} To overcome the scarcity of real-world robot data, recent efforts turn to leverage robot simulation~\cite{simulation_data:robotwin1,simulation_data:libero} or egocentric human videos~\cite{vla:egovla,human_data:egodex,vla:hrdt}. By scaling up diverse manipulation tasks in simulation, RoboTwin series~\cite{simulation_data:robotwin1,simulation_data:robotwin2} have demonstrated improved generalization of VLAs for real-world robot control. Furthermore, easy-collected human videos, which can be viewed as a large-scale source of behavior priors, have been widely used to improve robot visual encoder for embodied vision~\cite{robot:HRAlign}, provide scalable source of supervision for VLA pre-training~\cite{vla:hrdt,vla:egovla}, and learning dexterous manipulation in post-training~\cite{robot-learn:ImMimic,robot-learn:human2locoman}. However, directly pre-training VLAs on simulation data or human data, suffers from Sim2Real gap and large discrepancies in action spaces. In contrast, we introduce human-robot mutual imitation pre-training with bidirectional action space conversion, successfully integrating robot behavior priors in simulation data and real-world human action knowledge into a unified model. Most importantly, our work demonstrates that it is possible to achieve a generalizable VLA without real-world robot data when a properly designed pre-training recipe is used.

\section{Proposed Approach}
\label{sec:method}
In this section, we present methodology of our model MiVLA, which integrates behavioral priors in simulated robot data and real-world action knowledge in human data into a unified model. It begins with the problem formulation, followed by details of model architecture. Then, we present procedure of proposed mutual imitation.

\subsection{Problem Formulation}
Formally, simulated robot demonstrations are denoted as $D_{r}=\{l_r, v_{r}, a_{r}\}$, where $l_r$ denotes language instruction, $v_{r}$ represents a sequence of robot views, and $a_{r}$ denotes corresponding robot actions. We further denote human demonstrations as $D_{h}=\{l_h, v_{h}, a_{h}\}$, consisting of language descriptions $l_h$, a sequence of human visual observations $v_{h}$ and human actions $a_{h}$ across $T$ time steps. In general, VLAs are pre-trained on $\{D_{r}, D_{h}\}$ for learning multi-modal mappings from current robot observations $O_r^{t}=\{l_r^t, v_r^t, s_r^t\}$ or human observations $O_h^{t}=\{l_h^t, v_h^t, s_h^t\}$ to next actions $\{A_r, A_h\}$:
\begin{equation}
	\begin{array}{lll}
		 P_\theta(A_r\mid O_r^t),	& P_\theta(A_{h}  \mid O_h^t)&\\ 
	\end{array}
	\label{equ.vla} 
\end{equation}
where $P_\theta(\cdot)$ is the robot policy with learnable parameters $\theta$. $s_r^t$ represents proprioception of a robot at $t$-th time step while $s_h^t$ is the counterpart for humans. $A_r=\{\bm{a}_r\}^{t+H}_{t}$ denotes a chunk of $H$ robot actions, and $A_h = \{\bm{a}_h\}^{t+H}_{t}$ denotes the human action trajectory.

To achieve a general-purpose robot policy, a vision-language-action model should satisfy three criteria: architectural scalability, action diversity, and generalization. The scalability indicates an ability to encode multi-modal observations. Action diversity means versatility of the model, making it suitable for various behaviors across heterogeneous embodiments. Generalization ensures a model robust enough to unseen conditions. Next, we present details of proposed model that embody these three attributes, collectively referred as MiVLA.

\subsection{Model Architecture}
As illustrated in Fig.~\ref{figure:framework}, we build MiVLA upon multi-modal tokenizers and diffusion-based action decoder.

\noindent\textbf{Observation tokenizers.} For tokenizing vision and language inputs, we adopt DINOv2~\cite{vfm:dinov2} and Siglip~\cite{vlm:siglip} as the vision tokenizers, and use T5~\cite{llm:t5} as the language tokenizer. Specifically, each image frame is tokenized at resolution of $224\times224$, resulting in 392 visual tokens in total. Furthermore, we adopt two MLPs to project vision and language tokens into a common embedding space with same dimension. To tokenize proprioceptive states, we follow the design choice of existing works\cite{vla:RDT-1,vla:pi0} and directly use three MLPs to encode them into a fixed set of vectors.  

\noindent\textbf{Action space.} To leverage complementary real-world behavioral knowledge from both simulated robot data and human videos, we build a unified action space upon embodiment-specific action space such as human joints and robot joints, and common action space like end-effector pose. The specification of this unified action space is listed below:
\begin{itemize}
	\itemsep0em 
	\item Human joints $\bm{a}_h^{\text{joint}}$ with 48 dimensions: bilateral wrist poses (3D position and 6D orientation for each hand, 18 dimensions in total) and fingertip positions (3D position for all fingers, 30 dimensions in total) 
	\item Robot joints $\bm{a}_r^{\text{joint}}$ with 14 dimensions: 6 joints with 1 gripper for each robotic arm.
	\item End-effector with 14 dimensions: 3D position, 4D quaternion for each embodiment.
\end{itemize} 

\noindent\textbf{Action decoder.} 
To enable continuous robot control, MiVLA uses a diffusion transformer as the action decoder, where each computational block is built upon self-attention/cross-attention mechanism. Based on this architecture, the action generation can be regarded as an iterative denoising procedure using flow-matching~\cite{diffusion:flow_matching}: first, a noisy action chunk with a size of $H$ is encoded into a vector using MLPs; then, the diffusion transformer takes as input noised actions while observation tokens (i.e., vision tokens, language tokens and proprioceptive state tokens) are adopted as conditions. Finally, it estimates noises which are added to actions and recovers pure actions thereby.

\subsection{Human-Robot Mutual Imitation}
In this work, we aim to learn general-purpose robot control policies from easily collected simulated robot data with human videos. To achieve this, we first bridge the gap between human-robot action spaces, and then leverage abundant simulated robot and human data to learn general behavioral priors within a unified model.

\subsubsection{Human-Robot Action Mapping}
To enhance the transferability of action data across different embodiments, we establish a general conversion mechanism for human-robot motion mappings. Specifically, the thumb knuckle pose in human action space and end-effector pose of robot are selected as the reference points. The rest of joints can be inferred from reference points via inverse kinematic or anatomical priors. Formally, we denote human thumb knuckle as $\{\bm{a}^{\text{l-thumb}}_h, \bm{a}^{\text{r-thumb}}_h\}$, and robot EEF pose as $\{\bm{a}^{\text{l-eef}}_m, \bm{a}^{\text{r-eef}}_m\}$. In the following, we present the details of human-robot action mappings.

\noindent\textbf{Human-to-Robot Action Mapping.} Given initial robot EEF pose, termed as $\{\bm{a}^{\text{l-eef}}_{m,0}, \bm{a}^{\text{r-eef}}_{m,0}\}$, the target pose at time step $t$ can be obtained by transforming relative variation of human wrist pose:
\begin{equation}
	\begin{aligned}
	\bm{a}^{\text{l-eef}}_{r,t} & = \bm{a}^{\text{l-eef}}_{r,0} + \bm{R^h}(\bm{a}^{\text{l-thumb}}_{h,t} - \bm{a}^{\text{l-thumb}}_{h,0}) \\
	\bm{a}^{\text{r-eef}}_{r,t} & = \bm{a}^{\text{r-eef}}_{r,0} + \bm{R^h}(\bm{a}^{\text{r-thumb}}_{h,t} - \bm{a}^{\text{r-thumb}}_{h,0}) \\	
	\bm{a}_{r,t}^{\text{l-joint}} & = \bm{f}_{IK}(\bm{a}^{\text{l-eef}}_{r,t}) \\
	\bm{a}_{r,t}^{\text{r-joint}} & = \bm{f}_{IK}(\bm{a}^{\text{r-eef}}_{r,t}) \\
	\end{aligned}
\end{equation}
where $\bm{R^h}$ is a rotation transformation matrix that converts the coordinate system of human actions into that of the robot. $\bm{f}_{IK}(\cdot)$ indicates an optimization-based inverse kinematics (IK) solver, which is implemented by PyBullet library\footnote{\url{https://pybullet.org}}. 

\noindent\textbf{Robot-to-Human Action Mapping.} To map robot action to human action, we adopt the robot EEF pose as the thumb knuckle pose. Furthermore, the human joints at time step $t$ are calculated:
\begin{equation}
	\begin{aligned}
		\bm{a}^{\text{l-thumb}}_{h,t} & = \bm{R^m}(\bm{a}^{\text{l-eef}}_{h,t}) \\
		\bm{a}^{\text{r-thumb}}_{h,t} & = \bm{R^m}(\bm{a}^{\text{r-eef}}_{h,t}) \\	
		\bm{a}_{h,t}^{\text{l-joint}} & = \bm{a}^{\text{l-thumb}}_{h,t} + \bm{f}_{d}(\bm{a}^{\text{l-thumb}}_{h,t}) \\
		\bm{a}_{h,t}^{\text{r-joint}} & = \bm{a}^{\text{r-thumb}}_{h,t} + \bm{f}_{d}(\bm{a}^{\text{r-thumb}}_{h,t})\\
	\end{aligned}
\end{equation}
where $\bm{R^m}$ is a rotation transformation matrix that converts the coordinate system of robot actions into that of the human. $\bm{f}_{d}(\cdot)$ indicates a estimation function that empirically outputs distances between thumb and fingers based on anatomical priors~\cite{article:hand_size}. 

Using the human-robot action conversion mechanism, we can produce complementary action data, whether from either human videos or robot demonstrations.

\subsubsection{Pre-training Objectives}
\label{sec:mutual_imitation}
Instead of learning multi-modal mappings from observations to embodiment-specific actions, we propose to learn cross-embodiment actions prediction from the observation of single embodiment:
\begin{equation}
	\begin{array}{lll}
		P_\theta(A_r, \hat{A}_{h}\mid O_r^t),	&	P_\theta(A_{h}, \hat{A}_{r}\mid O_h^t)&\\ 
	\end{array}
	\label{equ.mivla} 
\end{equation} 
where $\hat{A}_{h}$ is the predicted human actions from robot observations, and $\hat{A}_{r}$ represents predicted robot actions derived from human observations. Given robot demonstrations, the learning objective is to minimize square errors $\ell_r$ for learning robot action prediction and robot-to-human action imitation:
\begin{equation}
	\begin{array}{lll}
		\ell_{r2h}& =\|A_r-A_r^{\ast}\|_{2} + \|\hat{A}_{h}-\hat{A}_{h}^{\ast}\|_{2} &\\ 
	\end{array}
	\label{equ.r2h_loss} 
\end{equation} 
where $A_r^{\ast}$ is the labeled trajectories in robot demonstrations $D_r$, and $\hat{A}_{h}^{\ast}$ is the generated human actions which are produced by imitating robot demonstrations. Similarly, we utilize the $\ell_2$ loss to learn human action generation and human-to-robot imitation by using human demonstrations:
\begin{equation}
	\begin{array}{lll}
		\ell_{h2r}	& =\|A_h-A_h^{\ast}\|_{2} + \|\hat{A}_{r}-\hat{A}_{r}^{\ast}\|_{2} &\\ 
	\end{array}
	\label{equ.h2r_loss} 
\end{equation} 
where $A_h^{\ast}$ is the labeled actions in robot demonstrations $D_r$, and
$\hat{A}_{r}^{\ast}$ represents synthesized robot actions derived from the reverse imitation of human actions.

Given a batch of demonstrations, the overall pre-training objective combines the robot-to-human and human-to-robot losses:
\begin{equation}
	\begin{array}{lll}
	\mathcal{L}	& = \ell_{r2h}+ \ell_{h2r} &\\ 
	\end{array}
	\label{equ.mi_loss} 
\end{equation}

\section{Experiment}
\label{sec:experiment}

In this section, we seek to answer the following questions through comprehensive experiments: 
1) How does proposed MiVLA perform compared to existing VLAs across multiple benchmarks and emobodiments ?
2) How does the design choice in MiVLA contribute to robot learning performance ?
3) To what extent does mutual imitation enhance the generalization capability of VLAs ?

Next, we first present experimental setup. Then, we demonstrate the merit of MiVLA through a comprehensive comparison with state-of-the-arts on both simulation and real-robot tasks. Finally, we conduct ablation studies to illustrate major properties of proposed MiVLA.

\begin{figure}[t]
	\centering
	\includegraphics[width=0.99\linewidth]{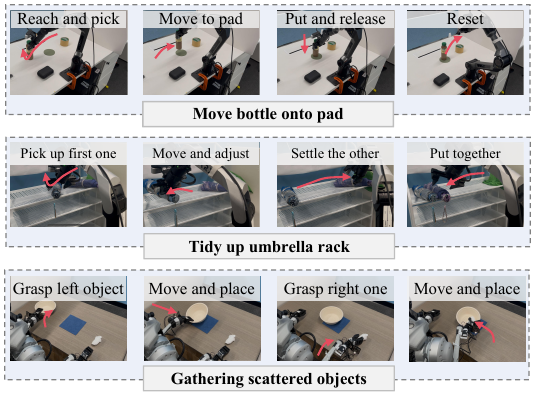}
	\caption{The overview of three designed tasks across three different robots. Red arrows indicate robot actions.}
	\label{fig:real-rotob-tasks}
\end{figure}

\begin{table*}[t]
	\centering
	\renewcommand\arraystretch{1.3}
	\resizebox{0.9\linewidth}{!}{
		\begin{tabular}{l|rr|rr|rr|rr|rr}
			\toprule
			\multicolumn{1}{c|}{\multirow{2}{*}{Tasks}}  & \multicolumn{2}{c|}{ACT} & \multicolumn{2}{c|}{$\boldsymbol{\pi}_{0}$}                             & \multicolumn{2}{c|}{$\boldsymbol{\pi}_{0.5}$}                          & \multicolumn{2}{c|}{H-RDT}                           & \multicolumn{2}{c}{MiVLA}     \\
			& Easy       & Hard       & Easy & Hard & Easy        & Hard     & Easy & Hard & Easy & Hard  \\
			\midrule
			Blocks Ranking Size                               &       0\%     &      0\%      & 0\%                      & 1\%                      &    3\%         &  17\%                           & 3\%                      & 3\%            & \textbf{21\%}      & \textbf{34\%}                \\
			Handover Block                                       &     0\%       &    1\%        & 1\%                      & 2\%                      &      12\%       & 22\%     & 3\%                      & 3\%                      & \textbf{66\%}  & \textbf{42\%}                    \\
			Hanging Mug                                   &     0\%       &      0\%      & 5\%                      & 3\%                      &    6\%         &   14\%                         & 5\%                      & 5\%            & \textbf{19\%}    & \textbf{25\%}                  \\
			Move Can Pot                                         &    0\%        &   0\%         & 21\%                     & 18\%                     &     42\%        &   50\%                             & 48\%                     & 34\%                   & \textbf{74\%}    & \textbf{68\%}                  \\
			Move Stapler Pad                  &        0\%    &    0\%        & 2\%                      & 5\%                      &    13\%         &   26\%                             & 4\%                      & 8\%              & \textbf{30\%}    & \textbf{33\%}                     \\
			Move Playingcard Away              &      2\%      &   0\%         & 30\%                     & 42\%                     &    33\%         &   \textbf{84\%}                           & 20\%                     & 49\%           &  \textbf{76\%}   & 79\%                   \\
			Place A2B Left                                     &      1\%      &   0\%         & 4\%                      & 4\%                      &  15\%           &    49\%                         & 16\%                     & 25\%             & \textbf{51\%} & \textbf{55\%}                     \\
			Place Object Basket                               &      0\%      &    1\%        & 29\%                     & 40\%                     &  22\%           &   57\%                         & 7\%                      & 40\%                     &  \textbf{71\%}   & \textbf{74\%}                   \\
			Stack Blocks Two                    &     0\%       &   1\%         & 17\%                     & 18\%                     & \textbf{33\%}            &     \textbf{68\%}                       & 2\%                      & 2\%                      &   23\%  & 7\%                    \\
			Stack Bowls Three                     &       0\%     &     0\%       & 22\%                     & 28\%                     &   54\%          &  62\%                          & 46\%                     & 60\%                     &   \textbf{79\%}  & \textbf{70\%}                     \\
			Put Bottles Dustbin                    &    0\%        &    0\%        & 7\%                      & 1\%                      &  13\%           &    8\%                          & 9\%                      & 2\%                      &      \textbf{47\%} & \textbf{33\%}                  \\
			Put\_Object\_Cabinet                     &   0\%         &   0\%         & 0\%                      & 0\%                      &   0\%          &    1\%          & 0\%                      & 0\%                      &     \textbf{68\%}  & \textbf{48\%}                    \\
			\hline
			Press\_Stapler                        &      22\%      &     21\%       & 67\%                     & 60\%                     &  70\%           &  71\%                              & 57\%                     & 65\%                     &      \textbf{78\%} & \textbf{85\%}               \\
			Open\_Microwave                           &  7\%          &    1\%        & 7\%                      & 12\%                     &  64\%           &  66\%                               & 74\%                     & 64\%                     &  \textbf{76\%}    & \textbf{79\%}                  \\
			Open\_Laptop                        &     6\%       &     3\%       & 33\%                     & 35\%                     & 80\%          &  96\%                              & 60\%                     & 78\%           &    \textbf{98\%}  & \textbf{99\%}                    \\
			Dump\_Bin\_Bigbin                    &     5\%       &     16\%       & 33\%                     & 49\%                     & 54\%            & 82\%                               & 95\%                     & 81\%             &     \textbf{98\%}   & \textbf{99\%}                   \\
			\hline
			Handover\_Mic                    &      10\%      &   10\%         & 16\%                     & 37\%                     & 45\%            &  89\%                              & 71\% & 94\%               &    \textbf{98\%}   & \textbf{99\%} \\
			Grab\_Roller                           &    33\%        &      60\%      & 60\%                     & 73\%                     &    63\%         &  99\%                              & 69\% & 80\%                     &  \textbf{100\%} & \textbf{100\%} \\
			Click\_Bell                       &    51\%        &    22\%        & 62\%                     & 55\%                     &     20\%        &  28\%                             & 75\% & 83\%                     &      \textbf{100\%}    & \textbf{99\%}                  \\
			Click\_Alarmclock                   &   36\%         &    22\%        & 53\%                     & 50\%                     &    57\%         &     62\%                        & 61\% & 74\%                     &    \textbf{100\%}  & \textbf{100\%}                    \\
			\hline
			Average (20 tasks)                            &      9\%      &   8\%         &  23\%   &     25\%     &     35\%        &    53\%         &   36\%  &  43\%   &  \textbf{69\%}    & \textbf{66\%} \\
			\bottomrule    
		\end{tabular}
	}
	\caption{Quantitative comparison with state-of-the-art robot policies across 20 simulated robot control tasks, with reported success rates on two evaluation modes.}
	\label{table:sota_sim_tasks}
\end{table*}
\begin{table*}[t]
	\centering
	\renewcommand\arraystretch{1.3}
	\resizebox{0.99\linewidth}{!}{
		\begin{tabular}{c|ccc|ccc|ccc|ccc|ccc}
			\toprule
			\multicolumn{1}{c|}{\multirow{2}{*}{\begin{tabular}[c]{@{}c@{}}Task \\ (embodiment) \end{tabular}}}  & \multicolumn{3}{c|}{ACT} & \multicolumn{3}{c|}{$\boldsymbol{\pi}_{0}$} & \multicolumn{3}{c|}{$\boldsymbol{\pi}_{0.5}$}  & \multicolumn{3}{c|}{H-RDT} & \multicolumn{3}{c}{MiVLA} \\
			& SR     & C     & T    & SR     & C     & T    & SR     & C     & T       & SR      & C     & T     & SR       & C      & T      \\
			\hline
			\begin{tabular}[c]{@{}c@{}}Move bottle onto pad \\ (PiPer) \end{tabular}                            &    0\%    &   0\%     &   -    &    20\%    &  37\%      &   66.2    &   \textbf{66\%}     &    \textbf{77\%}    &  56.2       &   36\%      &   47\%     &   \textbf{37.0}     &    54\%      & 71\%        &    37.2     \\
			\hline
			\begin{tabular}[c]{@{}c@{}}Tidy up umbrella rack\\ (ARX) \end{tabular}                       &   0\%         &  0\%      & -      &    60\%    &  70\%  &  70.9  & \textbf{75\%}      &  \textbf{85\%}      &   64.1                 & 45\%        &   55\%     &   60.9     &     60\%     &  70\%       &  \textbf{49.7}       \\
			\hline
			\begin{tabular}[c]{@{}c@{}} Gathering scattered objects \\ (LocoMan) \end{tabular}                                            &     0\%    &   0\%     &   -      &     \textbf{50\%}  &   60\%       &   26.0    &              20\%  &   30\%   &  43   &   0\%      &  0\%     &    -    &    \textbf{50\%}    &    \textbf{66\%}      &    \textbf{42.0}            \\
			\hline
			Average (3 tasks)       &        0\%    &   0\%     &   -   &     43.0\%  &   56\%     &   54.4     &   54\%            &  64\%      &   54.4     &    27\%     &   34\%      &   49.0     &     \textbf{55\%}    &    \textbf{69\%}    &  \textbf{43.0}      \\
			\bottomrule    
		\end{tabular}
	}
	\caption{Quantitative comparison with state-of-the-art robot policies across 3 real robot control tasks using 3 heterogeneous embodiments. The performance is evaluated using success rate (SR), completeness (C), and time cost (T), all of which are reported for each method.}
	\label{table:sota_real_tasks}
\end{table*}
\subsection{Experimental Setup}

We conduct comprehensive evaluations across both simulation and real-robot platforms:

\textbf{1) Simulation setup} We adopt RoboTwin-2.0 benchmark~\cite{simulation_data:robotwin2} with 50 dual-arm collaborative manipulation tasks for evaluation in simulation environment. It uses easy and hard modes to investigate generalization and robustness of VLAs, where the former offers clean environment and the latter further incorporates domain randomization with background changes and distracted objects. Following commonly-used protocol~\cite{vla:hrdt,simulation_data:robotwin2}, 2500 demonstrations (50 for each task) are selected for training. For fair evaluation, we adopt multi-task testing and apply VLAs to selected 20 representative tasks across three hardness levels, including 12 tasks at hard level, 4 at middle level and 4 at easy level.

\textbf{2) Real-robot setup} We conduct real-world robot manipulation tasks using three embodiments: unimanual AgileX PiPER, unimanual ARX-5 and bimanual LocoMan~\cite{robot-learn:human2locoman}. To ensure diversity of robot behaviors in real-world scenario and investigate cross-embodiment generalization, we design three robot control tasks:
\begin{itemize}
	\itemsep0em 
	\item \textbf{\textit{Move bottle onto pad}}: The robot must pick up a bottle from a random initial position, move it upon a pad, and place it precisely at a specified location. This task is performed using unimanual AgileX PiPER.
	\item \textbf{\textit{Tidy up umbrella rack}}: The robot's task is to tidy up a set of disordered umbrellas on a rack into a goal configuration. This is achieved by sequentially picking up each umbrella, rotating it to face the rack, and placing it back in a neat, left-to-right order. We adopt unimanual ARX-5 to conduct this task.
	\item \textbf{\textit{Gathering scattered objects}}: A robot must use bimanual manipulation to gather objects. This long-horizon task entails sequentially collecting and arranging them from largest to smallest. It is broken down into two substeps: first, pick up the large object and place it on the pad; then, pick up the small one and place it inside the large object. We conduct this task via bimanual LocoMan.  
\end{itemize} 
As illustrated in Fig.~\ref{fig:real-rotob-tasks}, these tasks are designed to evaluate not only the robot's ability to interact with external objects but also the capabilities for long-horizon collaborative control. For each task, we collect 30 demonstrations for post-training. During testing, we use three key metrics for comprehensive understanding of VLAs: (1) Success Rate (SR), the proportion of rollouts in which the task is successfully completed. (2) Completeness (C), the proportion of rollouts in which the sub-task is completed. (3) Time (T), the average number of seconds taken by the robot to complete a full task.  

\textbf{3) Training details} 
Proposed MiVLA is pre-trained on 4 A100 GPUs. We employ the AdamW optimizer with a batch size of 32 per GPU, resulting in a total effective batch size of 128. The initial learning rate is set to 1e-4 with a weight decay of 0.01, and a constant learning rate scheduler with a warmup phase is used. All pre-training experiments are conducted with \texttt{bf16} mixed precision to accelerate computation.
During the fine-tuning phase, we use 2 A100 GPUs with a per-GPU batch size of 16, for an effective batch size of 32. All optimizer parameters (e.g., learning rate, weight decay) remain consistent with that in pre-training stage. 

\textbf{4) Baselines} We evaluate MiVLA against five state-of-the-art baselines: ACT~\cite{robot-policy:act}, an action trunk transformer is trained from scratch for each task. $\boldsymbol{\pi}_{0}$~\cite{vla:pi0}, a state-of-the-art VLA is pre-trained on large-scale real-robot videos with over 10,000 hours episodes. $\boldsymbol{\pi}_{0.5}$~\cite{vla:pi05}, a recent improved version of $\boldsymbol{\pi}_{0}$ which is further pre-trained on open-world tasks in a variety of real homes. 
H-RDT~\cite{vla:hrdt}, a diffusion transformer pre-trained on EgoDex~\cite{human_data:egodex} for continuous robot control.
We use the open-sourced weights for baseline models and fine-tune them on the same datasets with same training configuration, ensuring the fairness in evaluation.

\subsection{Main Results}

\noindent\textbf{RoboTwin-2.0} As indicated in Table~\ref{table:sota_sim_tasks}, MiVLA achieves significant performance improvements over existing mainstream methods within both easy and hard mode. It achieves an average success rate of 69\% under easy mode and 66\% within hard mode, outperforming ACT (9\%, 8\%), $\boldsymbol{\pi}_{0}$ (23\%, 25\%), $\boldsymbol{\pi}_{0.5}$ (35\%, 53\%), H-RDT (36\%, 43\%). This strong performance especially within hard mode indicates that MiVLA exhibits superior manipulation capabilities robust to environment changes.

\noindent\textbf{Real-robot manipulation} Table~\ref{table:sota_real_tasks} lists real-world robot control performance for each method using three key metrics. It is found that proposed MiVLA achieves an average full-task success rate of 55\%, and sub-task success rate of 69\% , which are better than or competitive to existing methods across three robot platforms. Specifically, $\boldsymbol{\pi}_{0.5}$ achieves the best performance across two unimanual tasks while fail to perform bimanual task using LocoMan which is a composite embodiment. In a contrast, proposed MiVLA achieves the competitive results on two unimanual tasks, but performs best on bimanual task with composite embodiment. We conjecture there are two reasons behind this: 1) VLAs such as $\boldsymbol{\pi}_{0}$~\cite{vla:pi0} and $\boldsymbol{\pi}_{0.5}$~\cite{vla:pi05} are pre-trained on large-scale real robot data (over 10,000 hours) including PiPer and ARX, thereby exhibiting strong capability for adapting to unimanual tasks. 2) The LocoMan consists of quadrupedal robot and lightweight dual robotic arms, which is a unseen composite embodiment for pre-trained models. Despite difference, the MiVLA achieves an improved average success rates with mixed data at medium scale (around 900 hours, far less than that used for $\boldsymbol{\pi}$ series). This demonstrates that pre-training on simulated robot and human data with a proper training recipe can attain superior generalization capabilities comparable to that of VLAs pre-trained on real robot data.

\subsection{Ablation Study}
\noindent\textbf{The effect of pre-training choice.} Mutual imitation is the core of MiVLA, which involves two learning objectives, i.e., human-to-robot action imitation loss $\ell_{h2r}$ and robot-to-human action imitation loss $\ell_{r2h}$. We conduct a component-wise analysis by progressively using each learning objective for pre-training. Here, we adopt two baseline methods for investigation: 1) from scratch, the VLA model is directly trained on real robot control tasks; 2) human pre-train, where VLA model is pre-trained to learn human behaviors only without cross-embodiment imitation learning. As reported in Table~\ref{tab:abla_study_component}, directly training downstream real robot control tasks without pre-training performs worst, while pre-training on human data only brings limited performance. As progressively incorporates proposed cross-embodiment imitation learning objectives, the downstream performance of a VLA improves thereby. This illustrates that mutual imitation is particular beneficial for unlocking model's potential to learn general-purpose action knowledge from both simulated robot data and real-world human data.

\noindent\textbf{Few-shot adaptation.} We further perform few-shot adaptation to investigate the impact of each pre-training setting. As shown in Table~\ref{tab:abla_study_fewshot}, proposed mutual imitation is particularly beneficial for few-shot adaptation, where around 20 demonstrations is enough to adapt new tasks.

\begin{table}[t]
	\centering
	\renewcommand\arraystretch{1.2}
	\resizebox{0.99\linewidth}{!}{
	\begin{tabular}{l|c|ccc}
		\toprule
		\multicolumn{1}{c|}{Settings} & RobotWin2.0 & Piper & ARX & LocoMan \\
		\midrule
		From scratch               &      37\%       &     0\%               &         25\%             &      0\%              \\
		human pre-train        &    43\%         &     36\%               &       60\%               &    0\%                \\
		\hline
		$\ell_{h2r}$  &       46\%      &      30\%              &      49\%                &          20\%          \\
		$\ell_{h2r}$+$\ell_{r2h}$   &  66\%           &    54\%                &     60\%                 &   50\%                 \\
		\bottomrule               
	\end{tabular}
	}
	\caption{Investigating the effect of major components of MiVLA, including two simulation objectives.}
	\label{tab:abla_study_component}
\end{table}
\begin{figure*}[t]
	\centering
	\includegraphics[width=\linewidth]{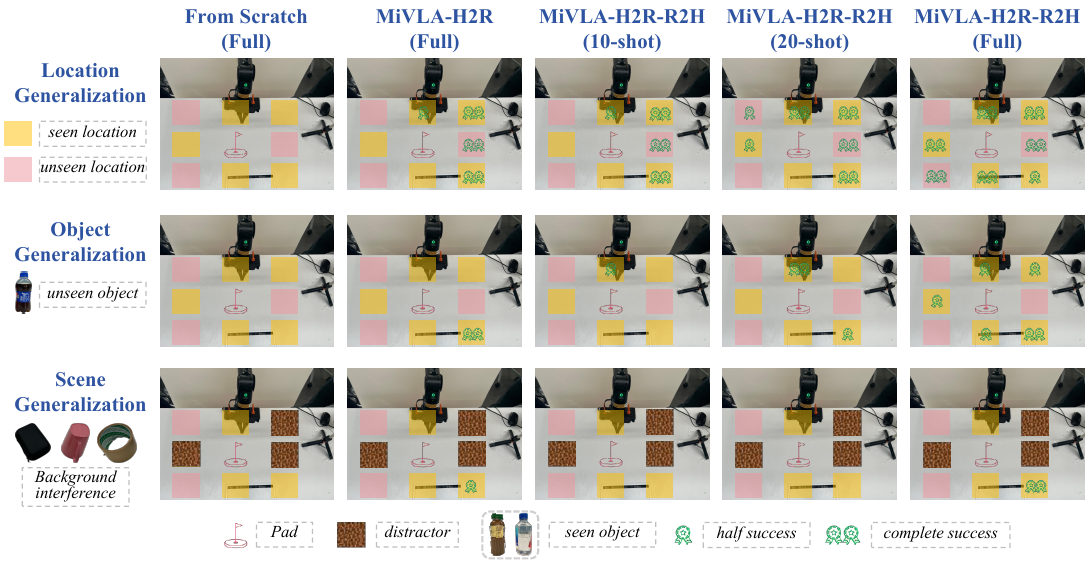}
	\caption{Qualitative investigation of generalization within three settings: cross-location generalization, cross-object generalization and cross-scene generalization.}
	\label{fig:visu}
\end{figure*}

\begin{table}[t]
	\centering
	\fontsize{9pt}{10pt}\selectfont
	\setlength{\tabcolsep}{3pt}
	\renewcommand\arraystretch{1.2}
	\resizebox{\linewidth}{!}{
		\begin{tabular}{>{\centering\arraybackslash}p{0.15\linewidth}|*{3}{>{\centering\arraybackslash}p{0.10\linewidth}}|*{3}{>{\centering\arraybackslash}p{0.10\linewidth}}}
			\toprule
			\multicolumn{1}{c|}{\multirow{2}{*}{Settings}} & \multicolumn{3}{c|}{Move bottle onto pad} & \multicolumn{3}{c}{Tidy up umbrella rack}\\
			\cline{2-7}
			\multicolumn{1}{l|}{} &  10 & 20 & \multicolumn{1}{c|}{30} & 5 & 10 & 20  \\
			\hline
			\multicolumn{1}{l|}{From scratch}              &      0\%       &     0\%               &  \multicolumn{1}{c|}{0\%}   &      10\%       &    10\%                &    25\%                     \\
			\multicolumn{1}{l|}{Human pre-train}    &      0\%       &   14\%     &  36\%   &    5\%         &   40\%                 &  45\%                     \\
			\multicolumn{1}{l|}{$\ell_{h2r}$}    &      12\%      &   26\%    & 30\%   &     35\%     &  50\%                 &  50\%                  \\
			\multicolumn{1}{l|}{$\ell_{h2r}$+$\ell_{r2h}$}         &      6\%       &     36\%     &  56\%  &     25\%        &     60\%               & 55\%  \\
			\bottomrule               
		\end{tabular}
	}
	\caption{Investigating impact of pre-training in few-shot adaptation.}
	\label{tab:abla_study_fewshot}
\end{table}

\begin{table}[t]
	\centering
	\fontsize{9pt}{10pt}\selectfont
	\setlength{\tabcolsep}{3pt}
	\renewcommand\arraystretch{1.2}
	\resizebox{\linewidth}{!}{
	\begin{tabular}{l|ccccc}
		\toprule
		Settings         & \begin{tabular}[c]{@{}c@{}}From scratch\\ (Full)\end{tabular} & \begin{tabular}[c]{@{}c@{}}MiVLA-H2R\\ (Full)\end{tabular} & \multicolumn{1}{c}{\begin{tabular}[c]{@{}c@{}}MiVLA\\ (10 shot)\end{tabular}} & \multicolumn{1}{c}{\begin{tabular}[c]{@{}c@{}}MiVLA\\ (20 shot)\end{tabular}} & \multicolumn{1}{c}{\begin{tabular}[c]{@{}c@{}}MiVLA\\ (Full)\end{tabular}} \\
		\hline
		Seen             & 0\%                                                           & 40\%                                                    & 10\%                                                          & 60\%                                                          & 75\%                                                       \\
		
		Unseen locations & 0\%                                                           & 33\%                                                    & 8\%                                                           & 33\%                                                          & 50\%                                                       \\
		
		Unseen objects   & 0\%                                                           & 20\%                                                    & 0\%                                                           & 10\%                                                          & 30\%                                                       \\
		
		Unseen scenes    & 0\%                                                           & 13\%                                                    & 0\%                                                           & 13\%                                                          & 38\%                                                       \\
		\hline
		Average          & 0\%                                                           & 30\%                                                    & 6\%                                                           & 36\%                                                          & 54\%                              \\
		\bottomrule                        
	\end{tabular}
}
\caption{Investigating impact of pre-training in generalization enhancement.}
\label{tab:abla_study_fewshot2}
\end{table}

\noindent\textbf{Generalization capabilities.} In this part, we seek to answer the question: to what extent does mutual imitation enhance generalization capabilities. As illustrated in Fig.~\ref{fig:visu} and Table~\ref{tab:abla_study_fewshot2}, we design three settings to investigate cross-location, cross-object, and cross-scene generalization for each pre-trained model. From the visualization results, three findings can be summarized: 1) Pre-training VLA with human-robot mutual imitation enhancing cross-location generalization, since human videos provide abundant behavior priors. 2) The benefits of mutual imitation in few-shot adaption manifest in two dimensions: cross-location and cross object generalization. 3) By using near 30 demonstrations, the MiVLA exhibits a godd generalization capability among cross-location, cross-object and cross-scenes settings.

\section{Conclusion} \label{Conclusion}
In this paper, we introduce MiVLA, a novel vision-language-action model empowered by mutual imitation learning. 
Specifically, we propose to learn generalizable VLA model by leveraging simulated robot data and human videos, since the former offers behavior priors for robot control while the latter provides extensive coverage of real-world tasks and scenes. 
Based on this mutual imitation, it integrates the behavioral fidelity of real-world human data with the manipulative diversity of simulated robot data into a unified model, thereby enhancing the generalization capability for downstream tasks. 
Extensive experiments conducted on both simulation and real-world platforms with three robots (ARX, PiPer and LocoMan), demonstrate that MiVLA achieves strong improved generalization capability, outperforming state-of-the-art VLAs More importantly, this study demonstrates a more accessible and cost-effective approach to developing generalizable VLAs without access real robot data, proving a scalable direction for developing generalizable VLAs.

\clearpage
\setcounter{page}{1}

\setcounter{figure}{0}
\setcounter{table}{0}
\renewcommand{\thefigure}{A\arabic{figure}}
\renewcommand{\thetable}{A\arabic{table}}

\begin{table*}[ht] 
	\centering %
	\begin{tabular}{@{}lcccccccccc@{}}
		\toprule
		& \multicolumn{2}{c}{ACT} & \multicolumn{2}{c}{$\Pi_{0}$} & \multicolumn{2}{c}{$\Pi_{0.5}$}      & \multicolumn{2}{c}{H-RDT} & \multicolumn{2}{c}{MiVLA}         \\ \midrule
		Task  Name                  & Easy       & Hard       & Easy          & Hard    & Easy          & Hard          & Easy        & Hard        & Easy            & Hard            \\
		blocks\_ranking\_rgb        & 0\%        & 0\%        & 1\%           & 5\%     & \textbf{17\%} & 42\%          & 3\%         & 2\%         & 13\%            & \textbf{47\%}   \\
		blocks\_ranking\_size       & 0\%        & 0\%        & 0\%           & 1\%     & 3\%           & 17\%          & 3\%         & 3\%         & \textbf{21\%}   & \textbf{34\%}   \\
		handover\_block             & 0\%        & 1\%        & 1\%           & 2\%     & 12\%          & 22\%          & 3\%         & 3\%         & \textbf{66\%}   & \textbf{42\%}   \\
		hanging\_mug                & 0\%        & 0\%        & 5\%           & 3\%     & 6\%           & 14\%          & 5\%         & 5\%         & \textbf{19\%}   & \textbf{25\%}   \\
		move\_can\_pot              & 0\%        & 0\%        & 21\%          & 18\%    & 42\%          & 50\%          & 48\%        & 34\%        & \textbf{74\%}   & \textbf{68\%}   \\
		move\_stapler\_pad          & 0\%        & 0\%        & 2\%           & 5\%     & 13\%          & 26\%          & 4\%         & 8\%         & \textbf{30\%}   & \textbf{33\%}   \\
		place\_a2b\_left            & 1\%        & 0\%        & 4\%           & 4\%     & 15\%          & 49\%          & 16\%        & 25\%        & \textbf{51\%}   & \textbf{55\%}   \\
		place\_object\_basket       & 0\%        & 1\%        & 29\%          & 40\%    & 22\%          & 57\%          & 7\%         & 40\%        & \textbf{71\%}   & \textbf{74\%}   \\
		stack\_blocks\_two          & 0\%        & 1\%        & 17\%          & 18\%    & \textbf{33\%} & \textbf{68\%} & 2\%         & 2\%         & 23\%            & 7\%             \\
		stack\_bowls\_three         & 0\%        & 0\%        & 22\%          & 28\%    & 54\%          & 62\%          & 46\%        & 60\%        & \textbf{79\%}   & \textbf{70\%}   \\
		put\_bottles\_dustbin       & 0\%        & 0\%        & 7\%           & 1\%     & 13\%          & 8\%           & 9\%         & 2\%         & \textbf{47\%}   & \textbf{33\%}   \\
		put\_object\_cabinet        & 0\%        & 0\%        & 0\%           & 0\%     & 0\%           & 1\%           & 0\%         & 0\%         & \textbf{20\%}   & \textbf{15\%}   \\
		press\_stapler              & 22\%       & 21\%       & 67\%          & 60\%    & 70\%          & 71\%          & 57\%        & 65\%        & \textbf{78\%}   & \textbf{85\%}   \\
		open\_microwave             & 7\%        & 1\%        & 7\%           & 12\%    & 64\%          & 66\%          & 74\%        & 64\%        & \textbf{76\%}   & \textbf{79\%}   \\
		move\_playingcard\_away     & 2\%        & 0\%        & 30\%          & 42\%    & 33\%          & 84\%          & 20\%        & 49\%        & \textbf{76\%}   & 79\%            \\
		open\_laptop                & 6\%        & 3\%        & 33\%          & 35\%    & 80\%          & 96\%          & 60\%        & 78\%        & \textbf{98\%}   & \textbf{99\%}   \\
		dump\_bin\_bigbin           & 5\%        & 16\%       & 33\%          & 49\%    & 54\%          & 82\%          & 95\%        & 81\%        & \textbf{98\%}   & \textbf{99\%}   \\
		handover\_mic               & 10\%       & 10\%       & 16\%          & 37\%    & 45\%          & 89\%          & 71\%        & 94\%        & \textbf{98\%}   & \textbf{99\%}   \\
		grab\_roller                & 33\%       & 60\%       & 60\%          & 73\%    & 63\%          & 99\%          & 69\%        & 80\%        & \textbf{100\%}  & \textbf{100\%}  \\
		click\_bell                 & 51\%       & 22\%       & 62\%          & 55\%    & 20\%          & 28\%          & 75\%        & 83\%        & \textbf{100\%}  & \textbf{99\%}   \\
		click\_alarmclock           & 36\%       & 22\%       & 53\%          & 50\%    & 57\%          & 62\%          & 61\%        & 74\%        & \textbf{100\%}  & \textbf{100\%}  \\
		adjust\_bottle              & 1\%        & 10\%       & 45\%          & 69\%    & 25\%          & 97\%          & 57\%        & 90\%        & \textbf{100\%}  & \textbf{96\%}   \\
		beat\_block\_hammer         & 0\%        & 5\%        & 44\%          & 35\%    & 69\%          & 64\%          & 25\%        & 35\%        & \textbf{95\%}   & \textbf{83\%}   \\
		lift\_pot                   & 0\%        & 16\%       & 5\%           & 8\%     & 54\%          & 84\%          & 27\%        & 31\%        & \textbf{100\%}  & \textbf{95\%}   \\
		blocks\_ranking\_rgb        & 0\%        & 0\%        & 1\%           & 5\%     & \textbf{17\%} & \textbf{42\%} & 0\%         & 0\%         & 0\%             & 2\%             \\
		move\_pillbottle\_pad       & 0\%        & 1\%        & 5\%           & 7\%     & 13\%          & 43\%          & 8\%         & 26\%        & \textbf{70\%}   & \textbf{71\%}   \\
		pick\_diverse\_bottles      & 3\%        & 0\%        & 14\%          & 10\%    & 31\%          & 44\%          & 17\%        & 20\%        & \textbf{59\%}   & \textbf{63\%}   \\
		pick\_dual\_bottles         & 5\%        & 2\%        & 13\%          & 17\%    & 40\%          & 34\%          & 4\%         & 26\%        & \textbf{58\%}   & \textbf{57\%}   \\
		place\_a2b\_right           & 1\%        & 0\%        & 2\%           & 6\%     & 25\%          & 41\%          & 10\%        & 28\%        & \textbf{62\%}   & \textbf{73\%}   \\
		place\_bread\_basket        & 1\%        & 0\%        & 10\%          & 14\%    & 9\%           & 43\%          & 5\%         & 29\%        & \textbf{56\%}   & \textbf{53\%}   \\
		place\_bread\_skillet       & 2\%        & 0\%        & 8\%           & 2\%     & 16\%          & 42\%          & 8\%         & 12\%        & \textbf{62\%}   & \textbf{51\%}   \\
		place\_burger\_fries        & 7\%        & 11\%       & 20\%          & 12\%    & 15\%          & 68\%          & 12\%        & 44\%        & \textbf{79\%}   & \textbf{83\%}   \\
		place\_can\_basket          & 1\%        & 1\%        & 4\%           & 6\%     & 11\%          & 40\%          & 14\%        & 31\%        & \textbf{41\%}   & \textbf{59\%}   \\
		place\_cans\_plasticbox     & 1\%        & 2\%        & 0\%           & 9\%     & 0\%           & 40\%          & 16\%        & 33\%        & \textbf{39\%}   & \textbf{56\%}   \\
		place\_dual\_shoes          & 1\%        & 0\%        & 6\%           & 8\%     & 17\%          & 34\%          & 1\%         & 7\%         & \textbf{28\%}   & \textbf{36\%}   \\
		place\_empty\_cup           & 3\%        & 4\%        & 17\%          & 32\%    & 57\%          & 87\%          & 29\%        & 67\%        & \textbf{90\%}   & \textbf{88\%}   \\
		place\_fan                  & 0\%        & 0\%           & 6\%           & 3\%     & 29\%          & 56\%          & 16\%        & 29\%        & \textbf{75\%}   & \textbf{75\%}   \\
		place\_mouse\_pad           & 0\%        & 0\%        & 2\%           & 4\%     & 3\%           & 16\%          & 0\%         & 10\%        & \textbf{24\%}   & \textbf{28\%}   \\
		place\_object\_scale        & 0\%        & 0\%        & 7\%           & 7\%     & 16\%          & 56\%          & 8\%         & 21\%        & \textbf{44\%}   & \textbf{63\%}   \\
		place\_object\_stand        & 0\%        & 1\%        & 25\%          & 25\%    & 56\%          & 75\%          & 22\%        & 42\%        & \textbf{61\%}   & \textbf{76\%}   \\
		place\_phone\_stand         & 2\%        & 0\%        & 6\%           & 5\%     & 31\%          & 53\%          & 13\%        & 19\%        & \textbf{65\%}   & \textbf{71\%}   \\
		place\_shoe                 & 0\%        & 0\%        & 19\%          & 30\%    & 35\%          & 68\%          & 18\%        & 31\%        & \textbf{89\%}   & \textbf{83\%}   \\
		rotate\_qrcode              & 2\%        & 0\%        & 5\%           & 17\%    & 52\%          & 66\%          & 39\%        & 71\%        & \textbf{84\%}   & \textbf{86\%}   \\
		scan\_object                & 0\%        & 0\%        & \textbf{1\%}  & 0\%     & \textbf{1\%}  & 0\%           & 0\%         & 0\%         & 0\%             & 0\%             \\
		shake\_bottle               & 27\%       & 20\%       & 89\%          & 73\%    & 93\%          & \textbf{98\%} & 87\%        & 84\%        & \textbf{99\%}   & \textbf{98\%}   \\
		shake\_bottle\_horizontally & 26\%       & 19\%       & 95\%          & 82\%    & 91\%          & \textbf{97\%} & 85\%        & 86\%        & \textbf{99\%}   & \textbf{97\%}   \\
		stack\_blocks\_three        & 0\%        & 0\%        & 1\%           & 2\%     & 12\%          & 34\%          & 0\%         & 0\%         & 1\%             & 0\%             \\
		stack\_bowls\_two           & 11\%       & 7\%        & 14\%          & 74\%    & 87\%          & \textbf{97\%} & 85\%        & 94\%        & \textbf{91\%}   & 93\%            \\
		stamp\_seal                 & 0\%        & 0\%        & 11\%          & 16\%    & 19\%          & \textbf{43\%} & 4\%         & 10\%        & \textbf{29\%}   & 31\%            \\
		turn\_switch                & 1\%        & 6\%        & 9\%           & 19\%    & 38\%          & 35\%          & 34\%        & 31\%        & \textbf{62\%}   & \textbf{71\%}   \\ \midrule
		Average(All 50 tasks)       & 5.4\%      & 5.4\%      & 19.1\%        & 21.7\%  & 33.6\%        & 53.8\%        & 27.4\%      & 37.2\%      & \textbf{62.0\%} & \textbf{63.6\%}
		
	\end{tabular}
	\caption{Complete success rates of all methods on the 50 tasks of the RoboTwin-2.0 benchmark. 
		Results are reported for both ``Easy'' and ``Hard'' variations. 
		The best-performing method, MiVLA, is frequently highlighted in bold.}
	\label{tab:suppl_full_sim_results} 
\end{table*}

\section*{A. Details of Simulation Experiments}

This section aims to provide more comprehensive settings and results for the simulation experiments mentioned in the main text. We will detail the domain randomization parameters employed in the ``Hard'' mode of the RoboTwin-2.0 benchmark, and present the complete evaluation results across all 50 tasks for a more thorough analysis.

\subsection*{A.1 Domain Randomization Settings}

To evaluate the robustness and generalization capabilities of our model, we introduce domain randomization in the ``Hard'' mode by perturbing key properties of the environment to simulate real-world diversity. Based on our experimental setup, we primarily adopt the following three randomization strategies:
\begin{itemize}
	\item \textbf{Visual Background and Texture Randomization:} The background of the simulated environment is randomly selected and applied from a diverse texture library.
	
	\item \textbf{Table Clutter Randomization:} Distractor objects of various geometric shapes and colors are randomly placed in non-critical areas of the workspace to increase scene complexity and visual clutter.
	
	\item \textbf{Lighting Condition Randomization:} The position, color, and intensity of the scene's light sources are sampled within a predefined range to simulate different lighting conditions.
\end{itemize}
In addition, we introduce minor geometric perturbations to the environment, such as random variations in the workbench height within a range of $\pm$3 cm.

\subsection*{A.2  Full Results on Simulation}
\begin{figure}[t]
	\centering
	\includegraphics[width=0.9\linewidth]{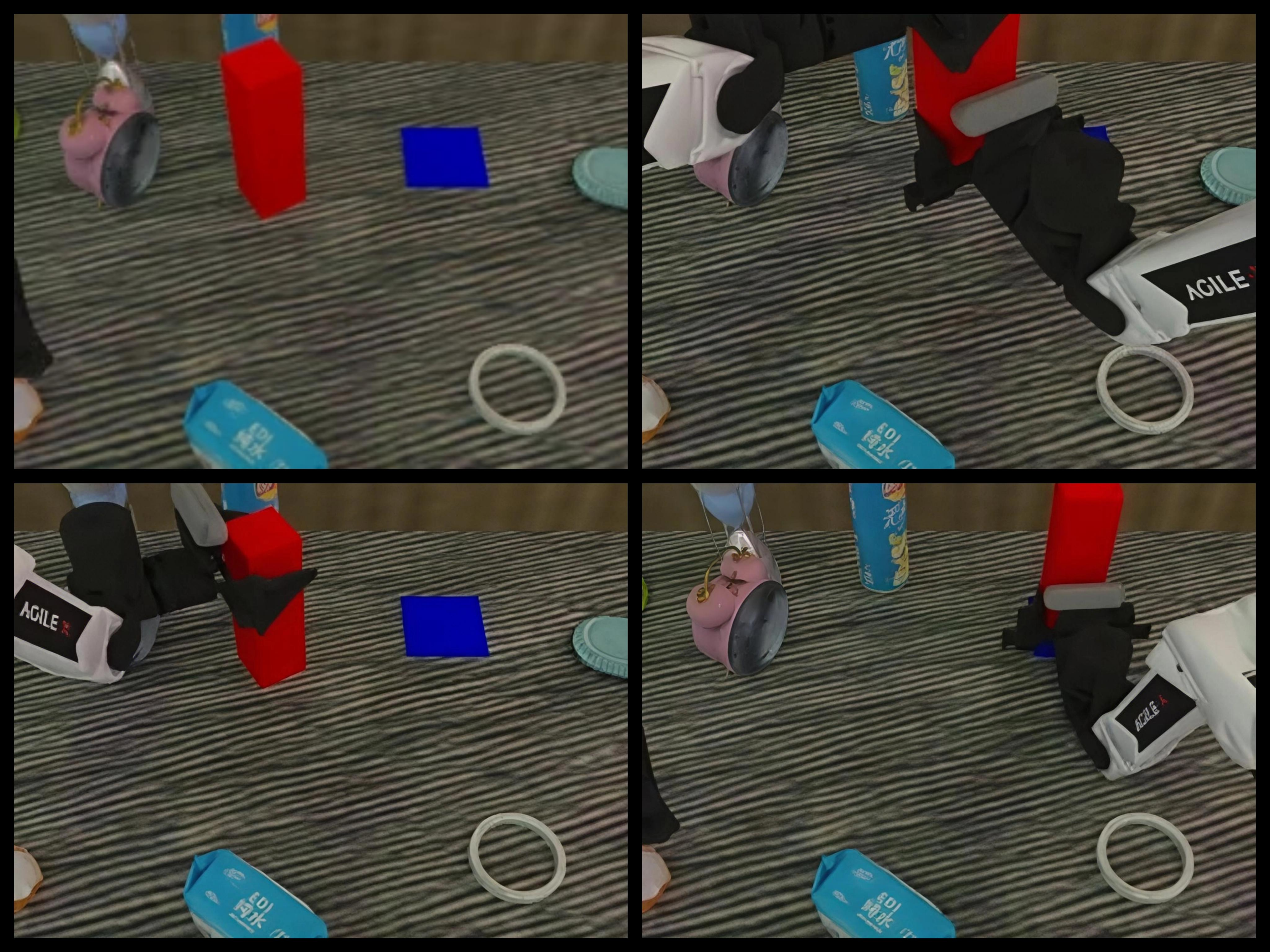} 
	\caption{An example of MiVLA's performance in the ``handover\_block" task within RoboTwin}
	\label{fig:robosim}
\end{figure}

\begin{figure}[t]
	\centering
	\includegraphics[width=0.9\linewidth]{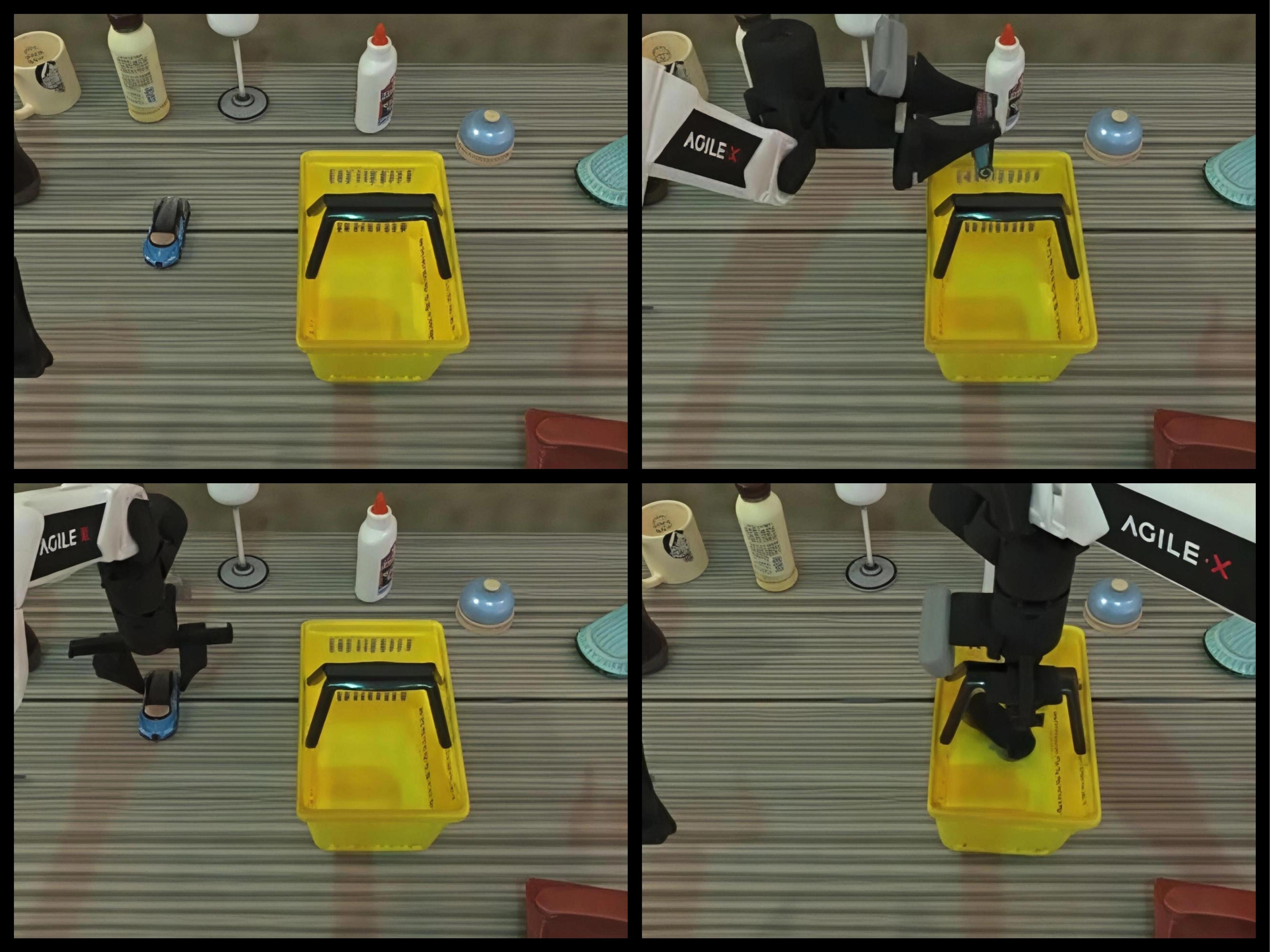} 
	\caption{An example of MiVLA's performance in the ``place\_object\_basket" task within RoboTwin}
	\label{fig:robosim2}
\end{figure}

This section presents the success rates (SR) of all baseline models across the entire suite of 50 tasks in the RoboTwin-2.0 benchmark.Figures~\ref{fig:robosim} and~\ref{fig:robosim2} showcase examples of Mivla's successful trajectories in representative tasks.The complete evaluation results are presented in Table~\ref{tab:suppl_full_sim_results}.These data further corroborate the conclusions drawn in the main text: our proposed MiVLA not only excels on the representative subset of tasks but also maintains a comprehensive performance lead across the entire task suite.

\begin{figure}[t]
	\centering
	\includegraphics[width=0.9\linewidth]{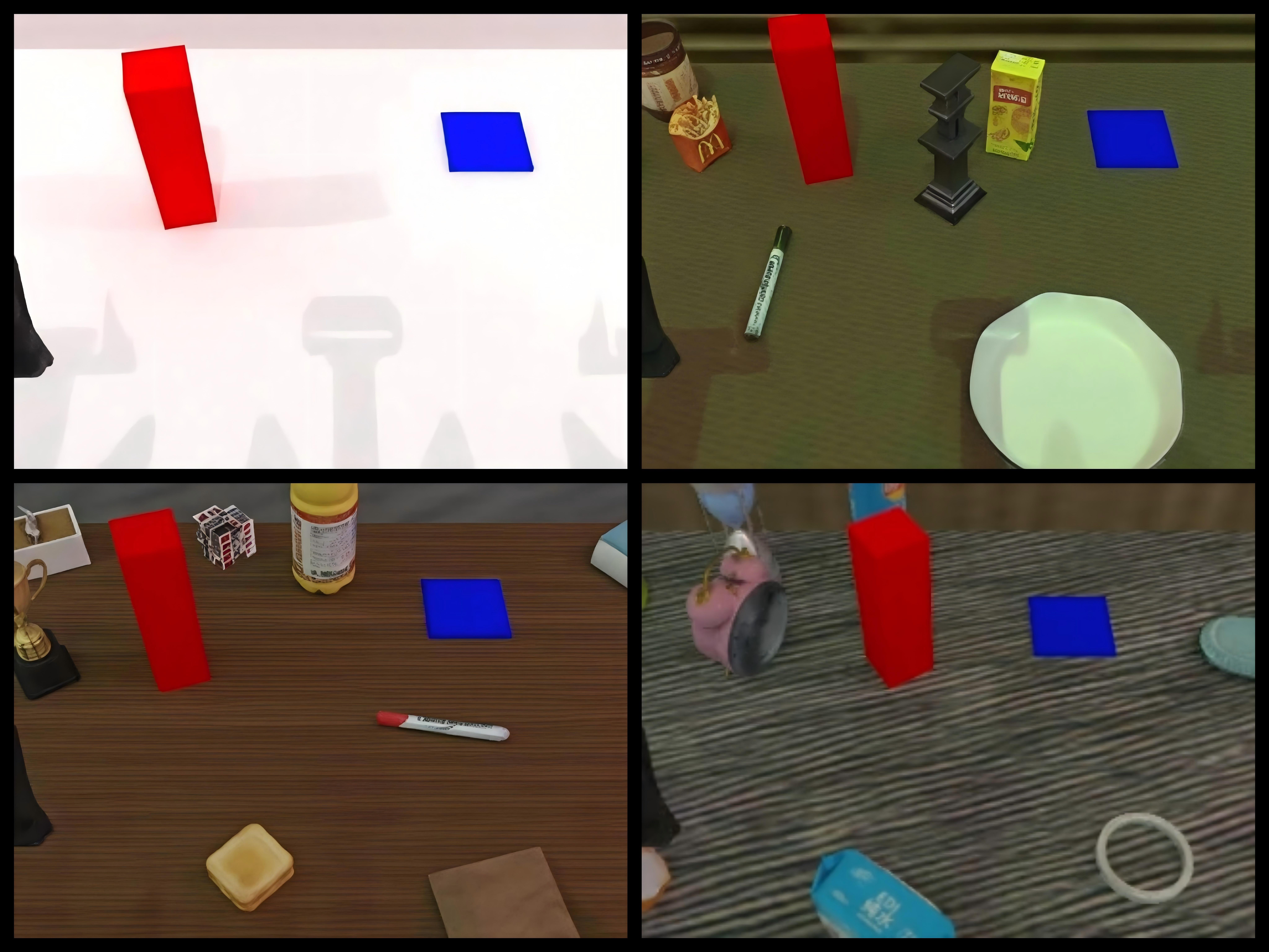} 
	\caption{Comparison of the `easy mode' (top-left) and `hard mode' (remaining images) environments in RoboTwin-2.0.}
	\label{fig:suppl_domain_rand}
\end{figure}

\section*{B. Details of Real-World Robot Experiment }

\subsection*{B.1 Robot Embodiment Specifications}

In this study, we employed three distinct robot embodiments with significant differences to rigorously evaluate the model's cross-embodiment generalization capabilities.

\begin{itemize}
	\item \textbf{AgileX PiPer \& ARX-5:} Both are table-top 6-DoF single-arm manipulators. Although they share the same number of degrees of freedom, they exhibit significant differences in their joint ranges, dynamic properties, and rotational joint characteristics.
	\item \textbf{LocoMan:} A composite robot embodiment composed of a quadruped robot and a lightweight dual-arm system. Its unique 6-DoF manipulation capability is provided by a hybrid combination of two parts: 
	(1)The first three DoF are realized through the leg movements of the quadruped; (2)The last three DoF are provided by three Dynamixel servos mounted on the robot's front legs. This hybrid-driven kinematic structure serves as a challenging test case to evaluate whether our VLA model can generalize knowledge learned from standard robot morphologies and adapt to a novel embodiment with a disparate structure.
\end{itemize}

\subsection*{B.2 Data Collection}
\addcontentsline{toc}{subsection}{B.2. Data Collection} 

To fine-tune the model, 30 successful expert demonstration trajectories were collected for every real-world task using two teleoperation methods. The AgileX PiPER and ARX-5 were controlled via a Leader-Follower scheme, where demonstrations were generated by an operator manipulating a kinematically identical master arm. The LocoMan was operated using a human pose-based solution, in which an operator's head and hand poses, tracked by an Apple Vision Pro, were mapped in real-time to the robot's base locomotion and dual-arm commands.

\subsection*{B.3 Qualitative Results and Analysis}

\begin{figure*}[t]
	\centering
	\includegraphics[width=0.9\textwidth]{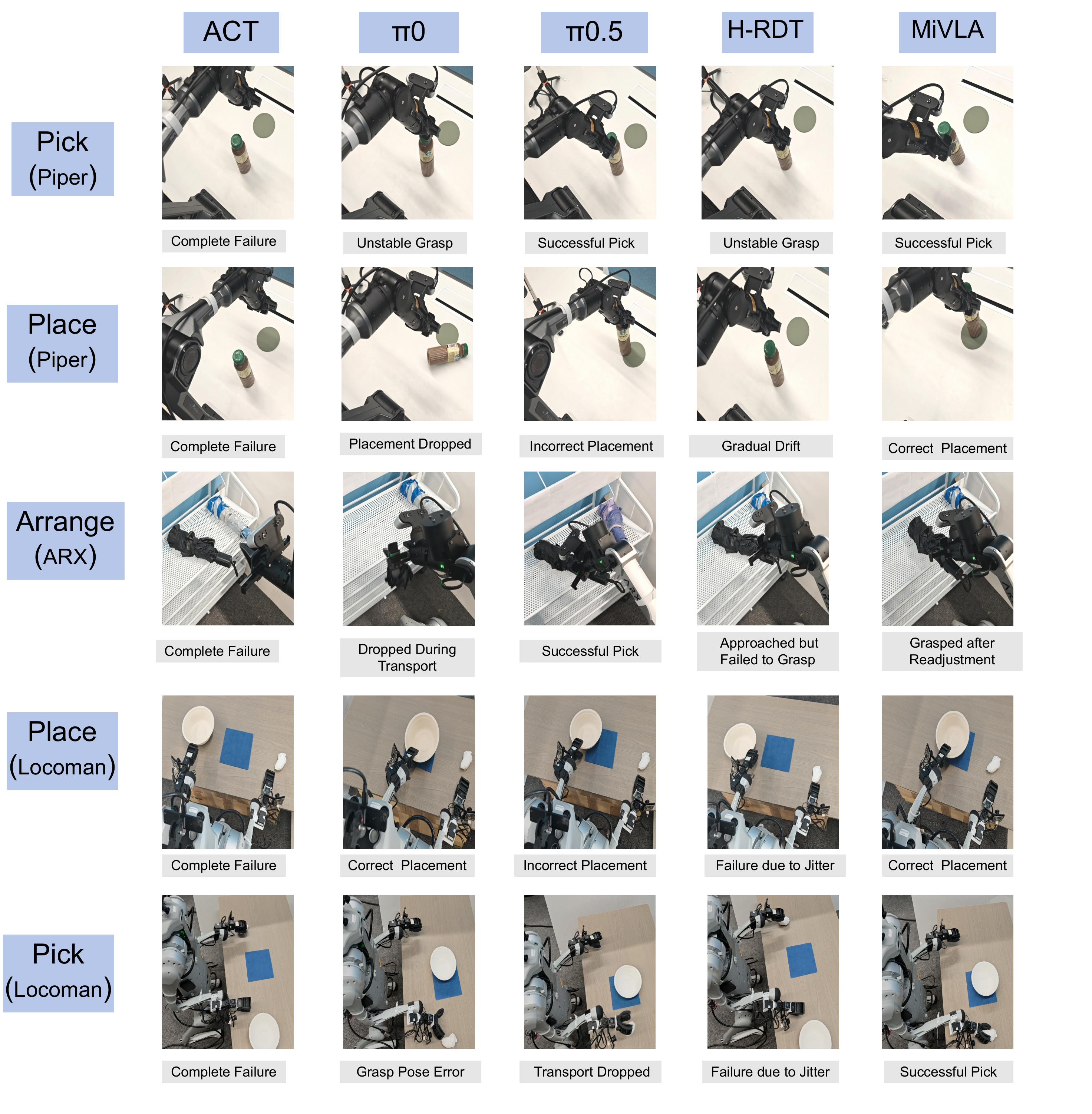} 
	\caption{ Qualitative comparison of policy performance on various real-world tasks and robot embodiments.}
	\label{fig:real_robot_comp}
\end{figure*}

To supplement the quantitative indicators presented in the experimental section of the text, this section provides a qualitative visualization of the final results of policy implementation in several representative practical tasks. Figure~\ref{fig:real_robot_comp} shows the comparison of the performance of MiVLA and all benchmark models in several representative scenarios. These visual results offer a deeper insight into the behavior of the model. From the observations, it can be seen that the baseline methods often exhibit some common failure modes, such as inaccurate grasping, item dropping, or inability to reach the target state. In contrast, The MiVLA model demonstrates higher accuracy and stronger stability, and is able to successfully complete tasks in all different robot forms. This not only highlights its effectiveness in terms of success rate, but also reflects the quality and time of the executed trajectories.

\section*{C. Limitations and Future Work}

Despite the strong performance of our MiVLA model, it is important to acknowledge its limitations, which primarily surface in out-of-distribution (OOD) scenarios. We identify three representative failure modes when the model encounters novel objects, unseen initial poses, and distracting backgrounds

\begin{figure}[t]
	\centering
	\includegraphics[width=0.3\textwidth]{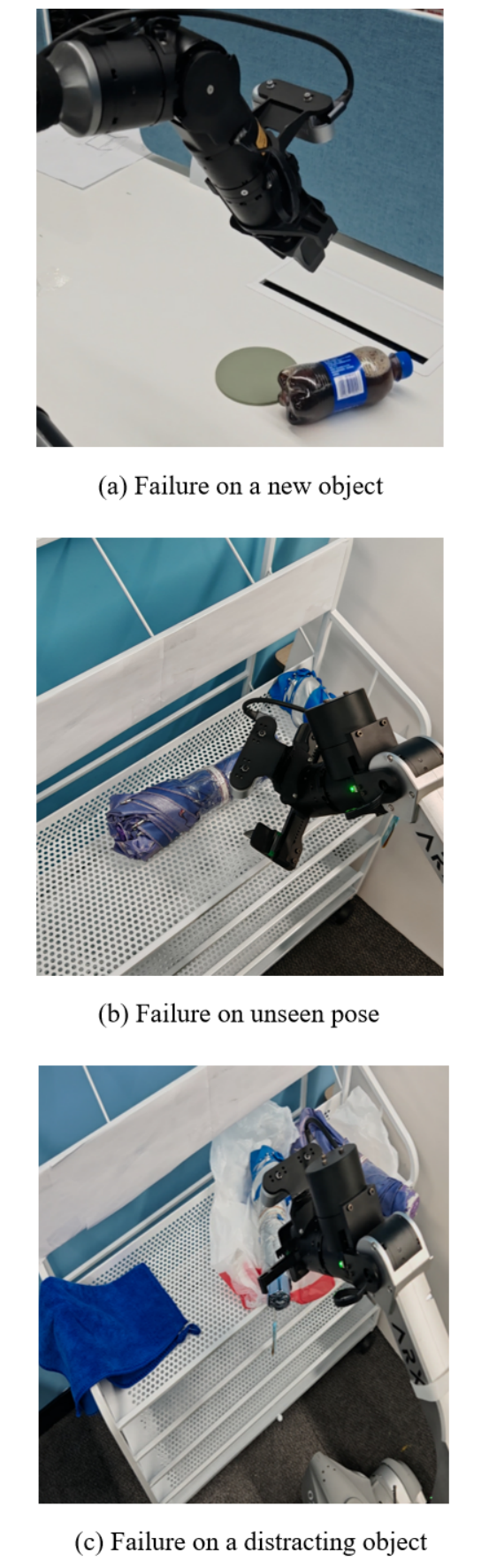} 
	\caption{An example of MiVLA's performance in the ``handover\_block" task within RoboTwin}
	\label{fig:limitations}
\end{figure}

\begin{itemize}
	\item \textbf{Novel Objects:} The model may struggle to generate appropriate grasping poses for objects with shapes and textures significantly different from the training data (Figure~\ref{fig:limitations}a).
	\item \textbf{Unseen Initial Poses:} When objects are placed in highly unusual or cluttered initial positions, the policy sometimes fails to find a valid trajectory, leading to collisions or inaction (Figure~\ref{fig:limitations}b).
	\item \textbf{Distracting Backgrounds:} Although trained with domain randomization, the model can still be distracted by highly complex or visually salient backgrounds that were not well-represented in the training distribution, causing it to misinterpret the task goal (Figure~\ref{fig:limitations}c).
\end{itemize}

It is important to note that while our MiVLA model demonstrates considerable generalization capabilities, 
its limitations become apparent in these more extreme or difficult out-of-distribution scenarios. 
We observe that the $\pi_{0.5}$ baseline exhibits better semantic generalization in some of these challenging cases. 
This superior performance can be attributed to its foundational architecture: 
$\pi_{0.5}$~\cite{vla:pi05} is built upon a Vision-Language Model (VLM) that was pretrained on a large-scale, multi-source, and heterogeneous dataset.

This highlights a fundamental trade-off between different architectural designs. 
Our MiVLA, based on a Diffusion Transformer, excels at directly learning complex visuo-motor policies from demonstration data. 
However, it lacks the explicit semantic and commonsense reasoning capabilities that VLMs acquire through their extensive pretraining. 
In essence, while our model masters the \textit{how} of a task through visual pattern recognition, 
models like $\pi_{0.5}$, owing to their VLM foundation, possess a better understanding of the what and why.

To address these limitations, a promising future direction is to integrate the cognitive reasoning abilities of VLMs 
with the powerful generative capabilities of diffusion-based policies. 
By leveraging a pretrained VLM to provide semantic guidance---such as object grounding, affordance prediction, or high-level planning---this 
fusion has the potential to enable the model to handle abstract language instructions, 
recover from errors through reasoning, and ultimately lead to more generalizable and human-like robotic intelligence.

{
    \small
    \bibliographystyle{ieeenat_fullname}
    \bibliography{main}
}


\end{document}